\documentclass[runningheads]{llncs}
\usepackage[T1]{fontenc}
\usepackage{graphicx}
\usepackage{amsmath,amsxtra,amssymb,latexsym, amscd}
\usepackage{times}
\usepackage{latexsym}
\usepackage{graphicx}
\usepackage{multirow}
\usepackage{pgf-pie}   
\usepackage{xcolor} 
\usepackage{tablefootnote}
\usepackage{tabularx}
\usepackage{adjustbox}
\usepackage{float}
\usepackage{graphicx}
\usepackage{caption}
\usepackage{lipsum}
\usepackage{flushend}
\usepackage{longtable}
\usepackage{makecell}
\usepackage{array,xcolor}
\usepackage{soul,soulutf8}
\usepackage{subcaption}

\usepackage[mathscr]{eucal}

\usepackage[T5]{fontenc}
\usepackage{CJKutf8}
\usepackage{hyperref}

\usepackage[style=numeric-comp, sorting=none, maxbibnames=99]{biblatex}
\begin{document}
\title{An Attempt to Develop a Neural Parser based on Simplified Head-Driven Phrase Structure Grammar on Vietnamese}
\titlerunning{An Attempt to Develop a Neural Parser based on Simplified HPSG on Vietnamese}
%
\author{Duc-Vu Nguyen\inst{2,3} \and
Thang Chau Phan\inst{1,3} \and
Quoc-Nam Nguyen\inst{1,3} \and
Kiet Van Nguyen\inst{1,3} \and
Ngan Luu-Thuy Nguyen\inst{1,3,}\thanks{Corresponding author}
}
\authorrunning{Duc-Vu Nguyen et al.}
%
\institute{Faculty of Information Science and Engineering, University of Information Technology, Ho Chi Minh City, Vietnam \and Laboratory of Multimedia Communications, University of Information Technology, Ho Chi Minh City, Vietnam \and
Vietnam National University, Ho Chi Minh City, Vietnam
\email{\{vund,kietnv,ngannlt\}@uit.edu.vn~\{20520929,20520644\}@gm.uit.edu.vn}}
\maketitle              
\begin{abstract}
In this paper, we aimed to develop a neural parser for Vietnamese based on simplified Head-Driven Phrase Structure Grammar (HPSG). The existing corpora, VietTreebank and VnDT, had around 15\% of constituency and dependency tree pairs that did not adhere to simplified HPSG rules. To attempt to address the issue of the corpora not adhering to simplified HPSG rules, we randomly permuted samples from the training and development sets to make them compliant with simplified HPSG. We then modified the first simplified HPSG Neural Parser for the Penn Treebank by replacing it with the PhoBERT or XLM-RoBERTa models, which can encode Vietnamese texts. We conducted experiments on our modified VietTreebank and VnDT corpora. Our extensive experiments showed that the simplified HPSG Neural Parser achieved a new state-of-the-art F-score of 82\% for constituency parsing when using the same predicted part-of-speech (POS) tags as the self-attentive constituency parser. Additionally, it outperformed previous studies in dependency parsing with a higher Unlabeled Attachment Score (UAS). However, our parser obtained lower Labeled Attachment Score (LAS) scores likely due to our focus on arc permutation without changing the original labels, as we did not consult with a linguistic expert. Lastly, the research findings of this paper suggest that simplified HPSG should be given more attention to linguistic expert when developing treebanks for Vietnamese natural language processing.
\end{abstract}

\section{Introduction}

Natural Language Processing (NLP) has witnessed significant advancements in recent years, propelled by the development of sophisticated models and algorithms. A critical area within NLP is the development of efficient and accurate parsers, particularly for languages with limited computational resources, like Vietnamese. Vietnamese, characterized by its tonal nature, complex morphology, and unique syntactic structure, presents unique challenges for parsing technologies \cite{vndt, Nguyen2018}. This study aims to address these challenges by developing a Vietnamese neural parser using a simplified version of Head-Driven Phrase Structure Grammar (HPSG) \cite{PollardSag94, do2008implementing}.

Our approach involves addressing inconsistencies within the VietTreebank and VnDT corpora, which are pivotal for Vietnamese NLP \cite{nguyen-2018-bktreebank, luong2013}. We incorporated advanced text encoding models, PhoBERT and XLM-RoBERTa, hypothesizing that these models would enhance the parser's performance due to their robust linguistic representation capabilities \cite{nguyen-tuan-nguyen-2020-phobert, conneau-etal-2020-unsupervised}. Our experiments demonstrate that the parser achieves an 82\% F-score in constituency parsing and shows promising performance in dependency parsing, outperforming others in the field despite lower scores in the Labeled Attachment Score (LAS) \cite{DozatM17, Chomsky+1993}.

In the context of the VLSP 2023 - Vietnamese Constituency Parsing Challenge, our study also ventures into transforming constituency trees into dependency trees using proposed head-rules implemented with the ClearNLP toolkit \cite{de-marneffe-etal-2006-generating, ma-etal-2010-dependency}. This transformation is particularly significant, as it was achieved without direct linguistic input. Remarkably, the HPSG Neural Parser \cite{zhou-zhao-2019-head} achieved a marginally higher F-score of 89.04\%, surpassing established parsers like the Stanza Constituency Parser, which scored 88.73\% \cite{qi-etal-2020-stanza}. This outcome underscores the potential of incorporating linguistic expertise into the development of Vietnamese NLP tools, an area that has been relatively underexplored.

The rest of this paper is structured in the following manner. Section \ref{related_work} surveys several current works on Vietnamese parsing. Section \ref{corpora} briefly looks and analyzes at the used datasets for our methodology and baseline. Our methodology is presented in detail through Section \ref{methodology}. Section \ref{experiments} illustrates processes for experimental settings, implementing models, and our experimental results on each dataset, and describes the result analysis and discussion of the proposed approach. In summary, Section \ref{conclusion} serves as the final part of our research and outlines our conclusions and any potential areas for future exploration.

\section{Background and Related Work}\label{related_work}

Vietnamese parsing has evolved significantly over the past decades. Early efforts were focused on building foundational resources like the VietTreebank and exploring basic dependency parsing strategies for Vietnamese. These efforts laid the groundwork for more sophisticated parsing techniques, as seen in \cite{vndt} and \cite{VLSP2023VietnameseParsing, nguyen-etal-2009-building}.

The role of treebanks in Vietnamese NLP cannot be overstated. \citeauthor{Nguyen2018} \cite{Nguyen2018} emphasized the importance of ensuring annotation consistency and accuracy in Vietnamese treebanks, which has been pivotal in advancing parsing techniques \cite{Nguyen2018}. These treebanks serve as critical resources for training and evaluating parsers, forming the backbone of most modern Vietnamese NLP applications, as demonstrated in studies like \cite{luong2013, nguyen-2018-bktreebank}.

Neural parsing techniques have seen significant innovations, shifting from traditional rule-based methods to more advanced neural network-based approaches. Key developments include the minimal span-based neural constituency parsers by \citeauthor{stern-etal-2017-minimal} \cite{stern-etal-2017-minimal} and the analytical insights into neural constituency parsers provided by \citeauthor{gaddy-etal-2018-whats} \cite{gaddy-etal-2018-whats}. These studies have significantly influenced the field, moving it towards more efficient and accurate parsing solutions.

The integration of pre-trained models like PhoBERT and XLM-RoBERTa into parsing has been a game-changer. \citeauthor{nguyen-tuan-nguyen-2020-phobert} \cite{nguyen-tuan-nguyen-2020-phobert} introduction of PhoBERT, and \citeauthor{conneau-etal-2020-unsupervised} \cite{conneau-etal-2020-unsupervised} work on unsupervised cross-lingual representation learning \cite{conneau-etal-2020-unsupervised} have demonstrated the potential of these models in enhancing parsing accuracy and efficiency, especially for languages like Vietnamese that lack extensive computational resources.

Despite these advancements, Vietnamese parsing faces specific challenges, such as the complexity of its syntactic structure and limited linguistic resources. Recent studies have proposed innovative solutions, including the use of head-rules for tree transformations and leveraging toolkits like ClearNLP to improve parsing efficiency \cite{vndt}. \citeauthor{8573351} \cite{8573351} presented a Dependency Tree-LSTM approach for Vietnamese sentiment analysis. \citeauthor{trang:hal-03329116} \cite{trang:hal-03329116} proposed a prosodic boundary prediction model to improve Vietnamese speech synthesis, using both traditional and novel features like syntactic blocks and links.

In comparing our parser's performance with others, such as the Stanza Constituency Parser, it becomes evident that while there are similarities in methodological approaches, each parser has its unique strengths and limitations. Our parser's slightly higher F-score highlights the potential impact of incorporating linguistic expertise in parser development, a concept that has been relatively underutilized in Vietnamese NLP \cite{linh-etal-2020-vlsp, kietsutag}.

The future of Vietnamese parsing looks promising, with potential impacts extending beyond the immediate field. The methodologies and findings from this study could influence future research directions, not only in Vietnamese NLP but also in the broader context of computational linguistics for other low-resource languages \cite{binheasy, nguyen-2019-neural}.

\section{Corpora}\label{corpora}
\subsection{VTB \& VnDT}

\citeauthor{nguyen2006lexicon} \cite{nguyen2006lexicon} introduce our project on developing a Vietnamese lexicon tailored for NLP applications, with a strong focus on the standardization of lexicon representation. Specifically, the authors propose a scalable framework of Vietnamese syntactic descriptions that are useful for defining tagsets and conducting morphosyntactic analysis.

The VietTreebank (VTB) dataset, developed by \citeauthor{nguyen-etal-2009-building} \cite{nguyen-etal-2009-building}, includes 20,000 sentences, with 10,000 having syntactic annotations and another 10,000 tagged for parts of speech. This dataset is sourced from ``Tuoi Tre\footnote{\url{https://tuoitre.vn/}}'', a Vietnamese newspaper. In addition, Nguyen et al. \cite{vndt} introduced a technique to convert Vietnamese treebanks into dependency trees, particularly important for addressing Vietnamese linguistic peculiarities. This conversion process led to the creation of the VnDT Treebank, featuring 10,200 sentences. The treebank was evaluated using two parsers, MSTParser and MaltParser, with MSTParser showing superior performance in Vietnamese dependency parsing. The VnDT Treebank is now a publicly available resource that offers significant value for research on Vietnamese natural language processing.

\begin{figure}[!h]
    \centering
    \includegraphics[width=\linewidth]{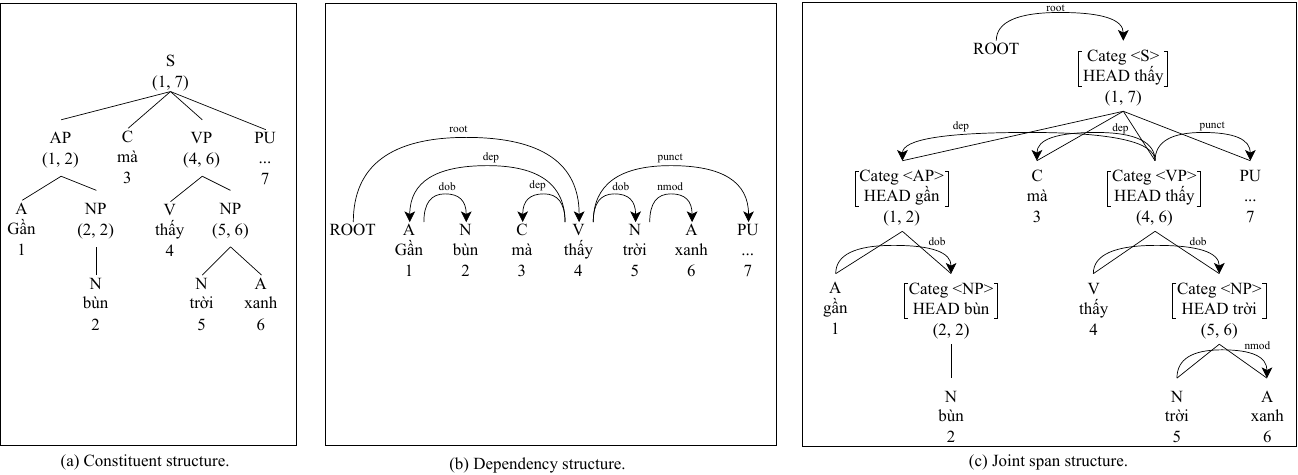}
    \caption{Constituent, dependency, and joint span structures, \textbf{\textit{extracted from the training datasets of VTB and VnDT and intended solely for visualization purposes, may contain slight labeling errors.}} These structures represent the same Vietnamese sentence, indexed from 1 to 7 and assigned an interval range for each node. The sentence ``Gần bùn mà thấy trời xanh'' translates to ``Close to the mud but seeing the blue sky.'' The Adjective Phrase (AP) ``Gần bùn'' corresponds to ``close to the mud,'' and the Verb Phrase (VP) ``thấy trời xanh'' corresponds to ``seeing the blue sky.'' Dependency arcs indicate grammatical relationships such as subject, object, and modifiers. The joint span structure combines constituent and dependency structures, explicitly marking the category (\textit{Categ}) and head word (\textit{HEAD}) for each span.}
    \label{fig:hpsg}
\end{figure}

Figure~\ref{fig:hpsg} presents an example \textbf{\textit{extracted from the training datasets of VTB and VnDT, which may contain slight labeling errors and is intended solely for visualization purposes,}} illustrating a valid simplified HPSG structure as proposed by \citeauthor{zhou-zhao-2019-head} \cite{zhou-zhao-2019-head}. This example highlights the integration of constituent and dependency analyses with explicit annotation of categories and head words. However, it was found that approximately 15\% of the constituency and dependency tree pairs in the VietTreebank and VnDT corpora did not adhere to the simplified HPSG framework outlined by \citeauthor{zhou-zhao-2019-head} \cite{zhou-zhao-2019-head}. To resolve this discrepancy, samples from the training and development sets were adjusted through random permutation to comply with these rules. Crucially, the original labels were preserved throughout the modification process, as no additional expert linguistic annotations were introduced to refine these corrections.

\subsection{VLSP 2023 Vietnamese Treebank}

The VLSP 2023 Vietnamese Treebank \cite{nguyen-etal-2009-building} is a collection of about 10,000 Vietnamese sentences, mostly from news articles and socio-political texts. The creators used various linguistic methods to handle language ambiguities, and annotators were assisted by automatic tools.

In the VLSP 2023 shared task \cite{VLSP2023VietnameseParsing}, participants are asked to develop a constituency parser. This parser takes a sentence and produces a tree that shows the grammatical structure of the sentence. Participants can improve their parsers using extra Vietnamese text or pre-trained language models. The evaluation uses Parseval metrics, and the test data includes texts from the same domain as the training data, as well as new areas like legal and biomedical fields.

\begin{figure}[H]
    \centering
    \begin{subfigure}[b]{0.48\linewidth}
        \includegraphics[width=\linewidth]{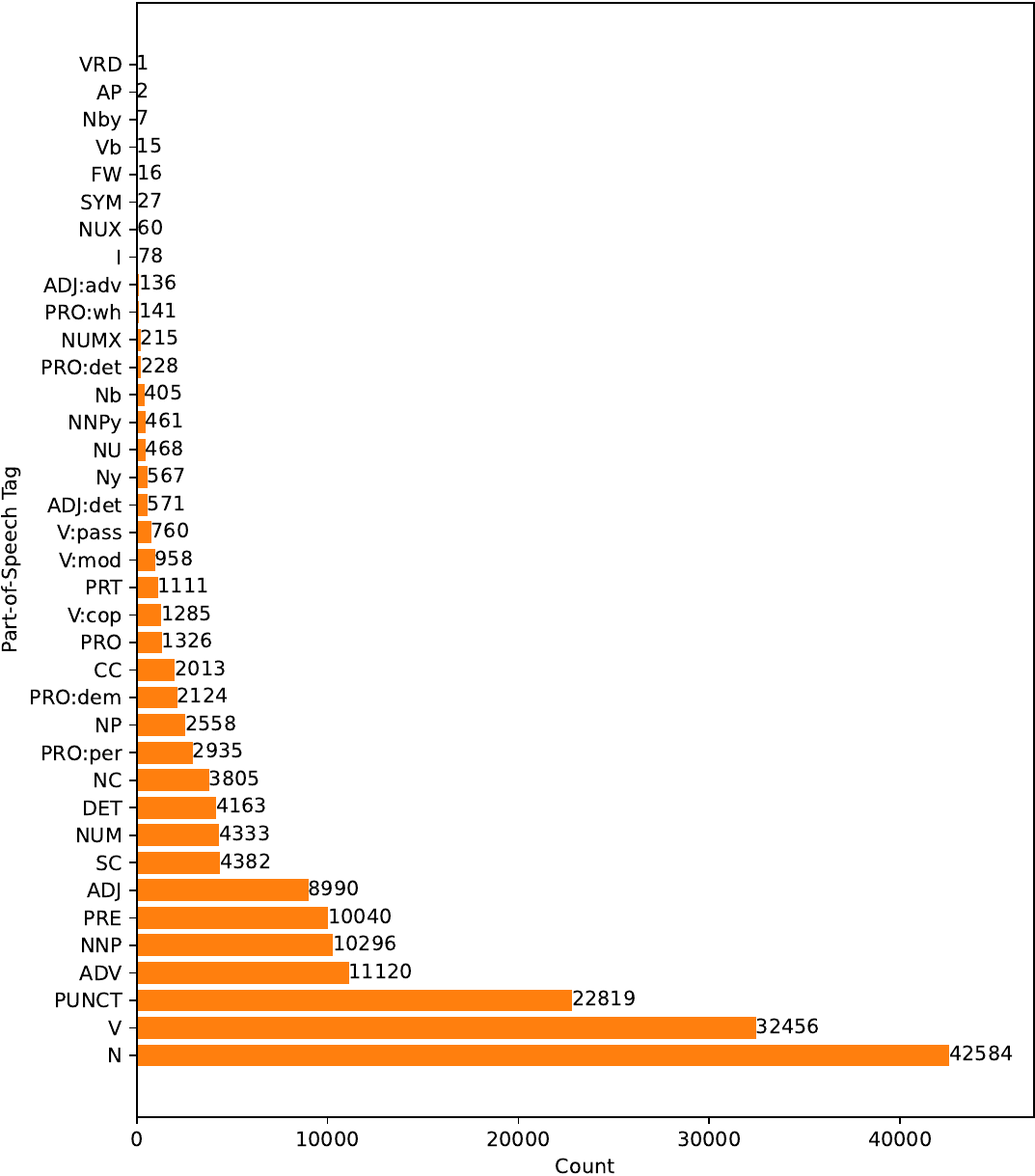}
        \caption{Part-of-speech tag distribution in the training set.}
        \label{fig:anlysis-viettreebank-pogtags}
    \end{subfigure}
    \hfill
    \begin{subfigure}[b]{0.48\linewidth}
        \includegraphics[width=\linewidth]{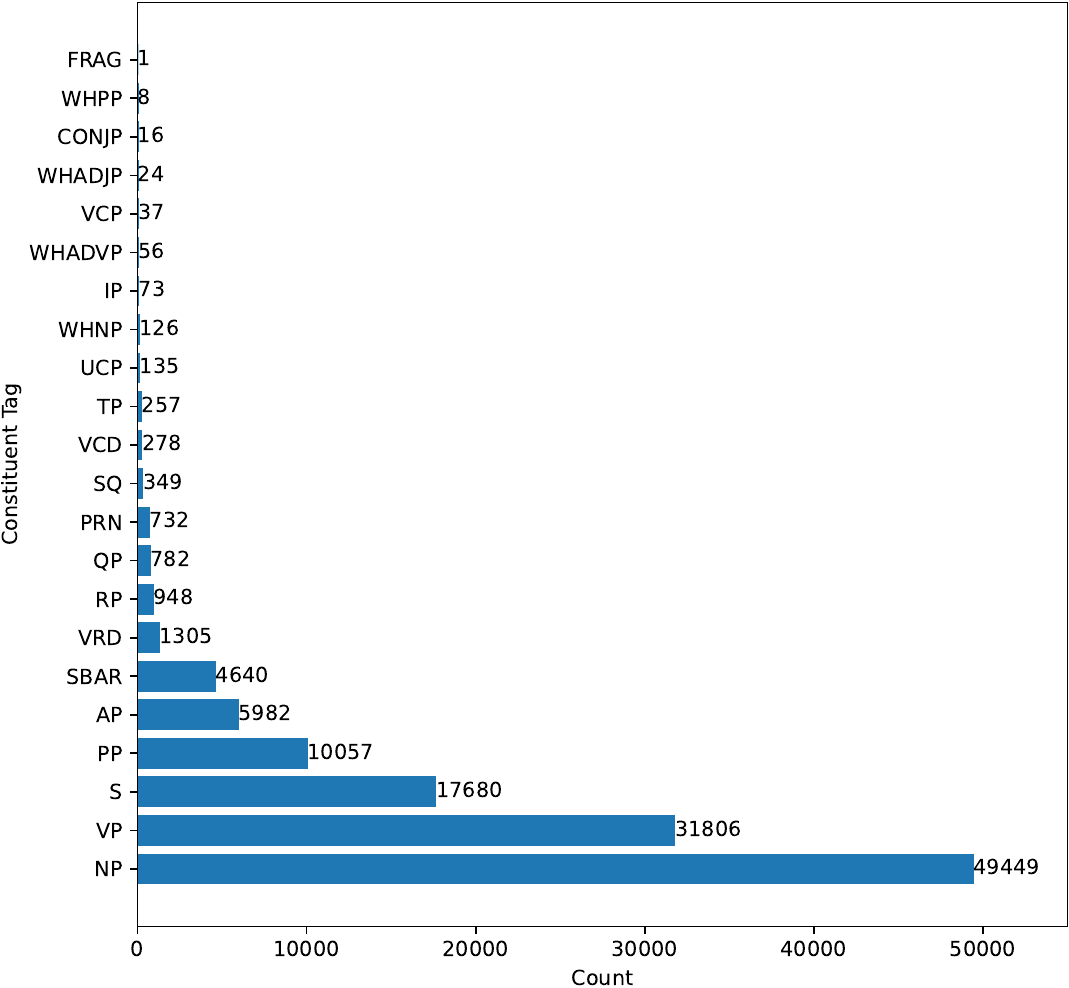}
        \caption{Constituent tag distribution in the training set.}
        \label{fig:viettreebank-constags}
    \end{subfigure}
    \caption{Distributions in the VLSP 2023 Vietnamese Treebank training set.}
    \label{fig:combined_viettreebank}
\end{figure}

We analyzed the training data from the VLSP 2023 Vietnamese Treebank to understand the structure of Vietnamese sentences. The data shows that nouns and verbs are the most common parts of speech, appearing 42,584 and 32,456 times respectively. This means that Vietnamese sentences in this dataset often focus on nouns and verbs, which is typical in formal writing like news articles.

Punctuation marks are also frequent, with 22,819 instances, highlighting the structured nature of written Vietnamese. The dataset includes various types of pronouns and verb forms, such as personal pronouns (\texttt{PRO:per}), demonstrative pronouns (\texttt{PRO:dem}), and copular verbs (\texttt{V:cop}), which are important features of the language.

In terms of sentence structure, noun phrases (\texttt{NP}) are the most common constituents, appearing 49,449 times. Verb phrases (\texttt{VP}) are also common, with 31,806 instances, indicating that sentences often have complex structures with detailed information. Other types of phrases like prepositional phrases (\texttt{PP}), adjective phrases (\texttt{AP}), and subordinate clauses (\texttt{SBAR}) are also present, showing the richness of Vietnamese syntax.

In summary, the VLSP 2023 Vietnamese Treebank provides valuable insights into the Vietnamese language, especially in formal writing. Understanding the common parts of speech and sentence structures helps in developing better natural language processing tools for Vietnamese.

\section{Method: Head-Driven Phrase Structure Grammar Neural Parser}\label{methodology}

\subsection{Overview of Joint Span HPSG}
Our HPSG Neural Parser is based on a novel Joint Span HPSG model, which innovatively integrates constituent and head information within a single constituent tree structure \cite{zhou-zhao-2019-head}. The ``joint span'' concept in this model encompasses all child phrases and dependency arcs among these phrases, providing a comprehensive syntactic analysis framework.

\subsection{Token Representation}
In this paper, we emphasize the importance of part-of-speech and contextual embeddings in the token representation of the HPSG Neural Parser. This parser's token representation is intricately designed, comprising character-level embeddings for morphological analysis, word embeddings to provide semantic context, and part-of-speech embeddings crucial for syntactic parsing \cite{DozatM17, nguyen-tuan-nguyen-2020-phobert}. The integration of these embeddings facilitates a comprehensive approach to token representation, establishing a robust foundation for accurate syntactic analysis. A notable enhancement to our parser is its integration with advanced pre-trained language models such as PhoBERT and XLM-RoBERTa \cite{nguyen-tuan-nguyen-2020-phobert, conneau-etal-2020-unsupervised}. This integration enables the parser to leverage deep, contextualized word representations, significantly improving its parsing performance, particularly in processing Vietnamese, a language with fewer linguistic resources.

\subsection{Self-Attention Encoder}
A key feature of our model is the self-attention encoder, based on the Transformer architecture \cite{vaswani-etal-2017-attention}. This encoder effectively contextualizes each word in a sentence, considering both immediate and distant word relations, thus capturing the complexity of syntactic structures in Vietnamese.

\subsection{Scoring Mechanism}
The scoring mechanism within the HPSG Neural Parser utilizes a biaffine attention model \cite{DozatM17}. This model accurately scores potential dependency relations among words, allowing for precise parsing and syntactic relationship establishment in complex sentences.

\subsection{Decoder for Joint Span HPSG}
The decoder in our HPSG Neural Parser employs dynamic programming to reconstruct syntactic parse trees from the scores generated by the encoder \cite{stern-etal-2017-minimal}. It utilizes an objective function that combines hinge loss and cross-entropy loss, optimizing the overall structure of the HPSG tree.

\section{Experiments and Results}\label{experiments}
\subsection{Baseline: Stanza Constituency Parser}

The Stanza Constituency Parser, as part of the Stanza open-source software distribution, has been implemented to support a wide range of languages, including Vietnamese. This general neural constituency parser, based on an in-order transition-based parsing framework, showcases its versatility and robustness \cite{qi-etal-2020-stanza, Bauer_Bui_Thai_Manning_2023}. In its application to the VLSP 2022 Vietnamese treebank, the parser achieved an impressive test score of 83.93\% F1, leading the private test leaderboard. This implementation uses a shift/reduce compiler-like mechanism, which manages a stack of partially constructed trees and a queue of unparsed words to predict transitions at each step of the parsing process. Integrating LSTM networks and attention mechanisms within this parser enhances its ability to parse syntactic structures accurately. In this study, the Stanza Constituency Parser is utilized as a benchmark to assess the performance of our HPSG Neural Parser.
\subsection{Experimental Settings}

\subsubsection{Stanza Part-of-Speech Tagger}

In our study, we utilized a modified version of the Stanza tagger\footnote{\url{https://github.com/stanfordnlp/stanza/blob/main/stanza/utils/training/run_pos.py}} \cite{qi-etal-2020-stanza}, fine-tuned with $\text{PhoBERT}_\text{large}$ \cite{nguyen-tuan-nguyen-2020-phobert}, achieving a 94\% token accuracy. This fine-tuning involved training an ensemble of 10 taggers on 90\% of the dataset, with the remaining 10\% used for validation, to generate the necessary tags for training our constituency model. For our training setup, we defined parameters including a cap of 15,000 steps, 12 BERT hidden layers, a learning rate of 1e-5, a batch size of 1,000, and the AdamW optimizer.

\subsubsection{Stanza Constituency Parser}

Our study involved refining the Stanza constituency parser\footnote{\url{https://github.com/stanfordnlp/stanza/blob/main/stanza/utils/training/run_constituency.py.py}} \cite{qi-etal-2020-stanza} with $\text{PhoBERT}_\text{large}$ \cite{nguyen-tuan-nguyen-2020-phobert}, yielding an 81\% F-score. We trained a solitary model, consisting of one parser, on 90\% of the data, allocating the remaining 10\% for validation. The training parameters were set with BERT fine-tuning from epoch 0 to 300, employing the AdamW optimizer, and a batch size of 32 for training.

\subsubsection{HPSG Neural Parser}

We discovered that around 1,000 constituency and dependency trees within the VTB and VnDT datasets failed to align with the HPSG tree criteria detailed in \cite{zhou-zhao-2019-head}. To rectify this, we applied a strategy of random permutation, which was not bound by linguistic constraints, to modify these trees to meet the specified HPSG criteria as cited in \cite{zhou-zhao-2019-head}. Notably, our modifications were limited to the training and development subsets of the VnDT dataset, while the test set remained unaltered. This approach ensures a fair comparison with previous studies conducted on the VnDT dataset.

We focused on enhancing the HPSG Neural Parser, which facilitates joint constituency and dependency decoding, as described in \cite{zhou-zhao-2019-head} and available\footnote{\url{https://github.com/DoodleJZ/HPSG-Neural-Parser}}. Our approach involved integrating $\text{PhoBERT}$ \cite{nguyen-tuan-nguyen-2020-phobert} and $\text{XLM-R}$ \cite{conneau-etal-2020-unsupervised} into the parser. We developed a single model that included parsers running for 100 epochs, utilizing both XLM-R and PhoBERT for the VTB \& VnDT datasets. However, for the VLSP 2023 Vietnamese Treebank, we exclusively used PhoBERT due to time constraints. The configuration of our parser included the use-tag feature, a learning rate of 0.00005, two layers of self-attention, and the AdamW optimizer.

\begin{figure}[!h]
    \centering
    \includegraphics[width=0.55\linewidth]{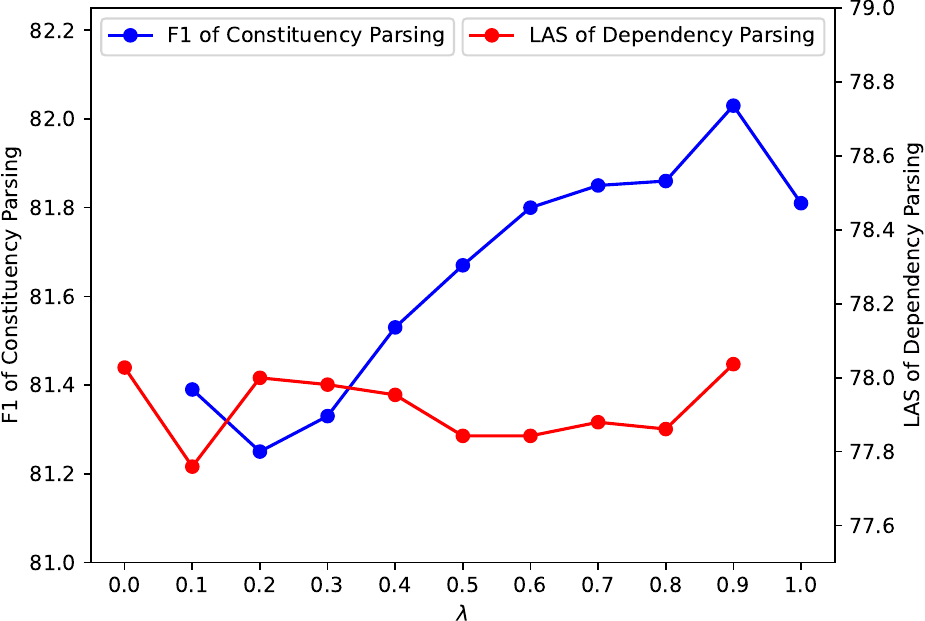}
    \caption{Balancing Constituency and Dependency in Joint Span HPSG Parsing on the VTB \& VnDT Development Sets.}
    \label{fig:balancing-vtb-vndt}
\end{figure}

In our study, we adjusted the hyper-parameter $\lambda$ within the HPSG Neural Parser framework. A higher $\lambda$ value assigns greater weight to the loss in constituency parsing relative to dependency parsing. As illustrated in Figure~\ref{fig:balancing-vtb-vndt}, a $\lambda$ setting of 0.9 yielded optimal results for both constituency and dependency parsing on the development datasets of VTB and VnDT. This outcome suggests that the impact of our approach, employing random permutation without linguistic constraints to align the trees with the HPSG criteria as mentioned in \cite{zhou-zhao-2019-head}, was minimal. Consequently, we adopted a $\lambda$ value of 0.9 for subsequent experiments in our research involving the HPSG Neural Parser.

\begin{figure}[!h]
    \centering
    \includegraphics[width=0.80\linewidth]{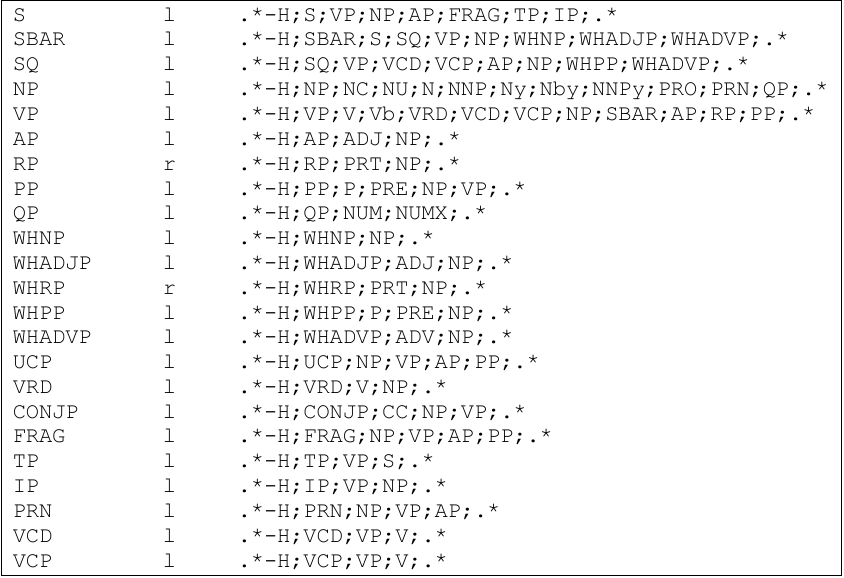}
    \caption{An attempted version of head rules for the VLSP 2023 Vietnamese Treebank was developed with a non-linguistic engineering background.}
    \label{fig:headrule}
\end{figure}

Currently, the VLSP 2023 Vietnamese Treebank comprises a constituency dataset annotated with head labels and does not include a corresponding dependency dataset. To adapt to this limitation and fulfill the requirements of the HPSG Neural Parser, we devised an initial set of head-percolation rules specifically for the VLSP 2023 Vietnamese Treebank. Notably, these rules were developed by individuals with a non-linguistic engineering background. Their design facilitates the conversion of constituency parsing to dependency parsing. This conversion process is demonstrated in Figure~\ref{fig:headrule}, providing a visual representation of the approach. For the practical application of this conversion, we modified a script from the ClearNLP framework\footnote{\url{https://github.com/clearnlp/clearnlp/tree/master/src/main/java/com/clearnlp/conversion}}.

\subsection{Results on VTB \& VnDT}

Table~\ref{tab:vtbvndt} presents comprehensive results from our experiments on the VTB and VnDT datasets, showcasing the proficiency of our parser in both constituency and dependency parsing. Notably, in constituency parsing, our parser achieved an impressive F-score of 82.34\%, indicating its exceptional performance. In dependency parsing, it outperformed other models in several metrics, though it achieved slightly lower Labeled Attachment Score (LAS) values. This outcome may be attributed to our emphasis on arc permutation without altering the original labels, as the adjustments were made without input from a linguistic expert.

\begin{table*}[h]
\caption{The performances of constituency parsing and dependency parsing on the VietTreebank and VnDT test sets were reported, respectively. The models were each run five times, as in previous studies. The predicted part-of-speech tags were generated using the VnCoreNLP toolkit \cite{vu-etal-2018-vncorenlp}. POS denoted part-of-speech.\label{tab:vtbvndt}}
\resizebox{\textwidth}{!}{%
\centering
\begin{tabular}{|clccc||clcc|}
\hline
\multicolumn{5}{|c||}{\textbf{Constituency Parsing}} & \multicolumn{4}{c|}{\textbf{Dependency Parsing}} \\ \hline
\multicolumn{2}{|c|}{\textbf{Model}} & \multicolumn{1}{c|}{\textbf{P}} & \multicolumn{1}{c|}{\textbf{R}} & \textbf{F} & \multicolumn{2}{c|}{\textbf{Model}} & \multicolumn{1}{c|}{\textbf{LAS}} & \textbf{UAS} \\ \hline\hline

\multicolumn{2}{|l|}{Self-Attentive w/ $\text{XLM-R}_\text{base}$ \cite{tran-2021-con}} & \multicolumn{1}{c|}{79.95} & \multicolumn{1}{c|}{78.61} & 79.28 & \multicolumn{2}{l|}{Biaffine w/ $\text{XLM-R}_\text{base}$ \cite{nguyen-tuan-nguyen-2020-phobert}} & \multicolumn{1}{c|}{76.46} & 83.10 \\ \hline

\multicolumn{2}{|l|}{Self-Attentive w/ $\text{XLM-R}_\text{large}$ \cite{tran-2021-con}} & \multicolumn{1}{c|}{80.78} & \multicolumn{1}{c|}{81.61} & 81.19 & \multicolumn{2}{l|}{Biaffine w/ $\text{XLM-R}_\text{large}$ \cite{nguyen-tuan-nguyen-2020-phobert}} & \multicolumn{1}{c|}{75.87} & 82.70 \\ \hline

\multicolumn{2}{|l|}{Self-Attentive w/ $\text{PhoBERT}_\text{base}$ \cite{tran-2021-con}} & \multicolumn{1}{c|}{81.14} & \multicolumn{1}{c|}{79.60} & 80.36 & \multicolumn{2}{l|}{Biaffine w/ $\text{PhoBERT}_\text{base}$ \cite{nguyen-tuan-nguyen-2020-phobert}} & \multicolumn{1}{c|}{\textbf{78.77}} & 85.22 \\ \hline

\multicolumn{2}{|l|}{Self-Attentive w/ $\text{PhoBERT}_\text{large}$ \cite{tran-2021-con}} & \multicolumn{1}{c|}{80.55} & \multicolumn{1}{c|}{80.54} & 80.55 & \multicolumn{2}{l|}{Biaffine w/ $\text{PhoBERT}_\text{large}$ \cite{nguyen-tuan-nguyen-2020-phobert}} & \multicolumn{1}{c|}{77.85} & 84.32 \\ \hline\hline

\multicolumn{1}{|c|}{\multirow{4}{*}{\begin{tabular}[c]{@{}c@{}}W/o\\ POS tags\end{tabular}}} & \multicolumn{1}{l|}{HPSG w/ $\text{XLM-R}_\text{base}$} & \multicolumn{1}{c|}{78.86} & \multicolumn{1}{c|}{79.06} & 78.96 & \multicolumn{1}{c|}{\multirow{4}{*}{\begin{tabular}[c]{@{}c@{}}W/o\\ POS tags\end{tabular}}} & \multicolumn{1}{l|}{HPSG w/ $\text{XLM-R}_\text{base}$} & \multicolumn{1}{c|}{75.64} & 83.51 \\ \cline{2-5} \cline{7-9} 
\multicolumn{1}{|c|}{} & \multicolumn{1}{l|}{HPSG w/ $\text{XLM-R}_\text{large}$} & \multicolumn{1}{c|}{80.98} & \multicolumn{1}{c|}{81.24} & 81.11 & \multicolumn{1}{c|}{} & \multicolumn{1}{l|}{HPSG w/ $\text{XLM-R}_\text{large}$} & \multicolumn{1}{c|}{77.67} & 85.06 \\ \cline{2-5} \cline{7-9} 
\multicolumn{1}{|c|}{} & \multicolumn{1}{l|}{HPSG w/ $\text{PhoBERT}_\text{base}$} & \multicolumn{1}{c|}{80.96} & \multicolumn{1}{c|}{81.76} & 81.36 & \multicolumn{1}{c|}{} & \multicolumn{1}{l|}{HPSG w/ $\text{PhoBERT}_\text{base}$} & \multicolumn{1}{c|}{77.60} & 85.18 \\ \cline{2-5} \cline{7-9} 
\multicolumn{1}{|c|}{} & \multicolumn{1}{l|}{HPSG w/ $\text{PhoBERT}_\text{large}$} & \multicolumn{1}{c|}{81.54} & \multicolumn{1}{c|}{82.32} & 81.93 & \multicolumn{1}{c|}{} & \multicolumn{1}{l|}{HPSG w/ $\text{PhoBERT}_\text{large}$} & \multicolumn{1}{c|}{78.16} & \textbf{85.73} \\ \hline
\multicolumn{1}{|c|}{\multirow{4}{*}{\begin{tabular}[c]{@{}c@{}}W/\\ POS tags\end{tabular}}} & \multicolumn{1}{l|}{HPSG w/ $\text{XLM-R}_\text{base}$} & \multicolumn{1}{c|}{79.49} & \multicolumn{1}{c|}{79.70} & 79.60 & \multicolumn{1}{c|}{\multirow{4}{*}{\begin{tabular}[c]{@{}c@{}}W/\\ POS tags\end{tabular}}} & \multicolumn{1}{l|}{HPSG w/ $\text{XLM-R}_\text{base}$} & \multicolumn{1}{c|}{75.98} & 83.51 \\ \cline{2-5} \cline{7-9} 
\multicolumn{1}{|c|}{} & \multicolumn{1}{l|}{HPSG w/ $\text{XLM-R}_\text{large}$} & \multicolumn{1}{c|}{81.57} & \multicolumn{1}{c|}{81.42} & 81.50 & \multicolumn{1}{c|}{} & \multicolumn{1}{l|}{HPSG w/ $\text{XLM-R}_\text{large}$} & \multicolumn{1}{c|}{77.92} & 85.14 \\ \cline{2-5} \cline{7-9} 
\multicolumn{1}{|c|}{} & \multicolumn{1}{l|}{HPSG w/ $\text{PhoBERT}_\text{base}$} & \multicolumn{1}{c|}{81.46} & \multicolumn{1}{c|}{82.11} & 81.78 & \multicolumn{1}{c|}{} & \multicolumn{1}{l|}{HPSG w/ $\text{PhoBERT}_\text{base}$} & \multicolumn{1}{c|}{77.92} & 85.36 \\ \cline{2-5} \cline{7-9} 
\multicolumn{1}{|c|}{} & \multicolumn{1}{l|}{HPSG w/ $\text{PhoBERT}_\text{large}$} & \multicolumn{1}{c|}{\textbf{82.03}} & \multicolumn{1}{c|}{\textbf{82.49}} & \textbf{82.34} & \multicolumn{1}{c|}{} & \multicolumn{1}{l|}{HPSG w/ $\text{PhoBERT}_\text{large}$} & \multicolumn{1}{c|}{78.42} & \textbf{85.73} \\ \hline
\end{tabular}}%
\end{table*}

In Table~\ref{tab:vtbvndt}, the enhanced performance of the models, especially those leveraging PhoBERT, is evident. The inclusion of part-of-speech (POS) tags significantly improved parsing accuracy, highlighting the critical role of POS information in these tasks. Lastly, the findings of this research underscore the need for greater consideration of simplified HPSG rules by linguistic experts when designing and refining treebanks for Vietnamese natural language processing. This comprehensive analysis provides valuable insights into the capabilities and limitations of different parsing models in addressing the complexities of the Vietnamese language.

\begin{table}[h]
\caption{The performances of dependency parsing on the VnDT test set were reported. [$\bigstar$] represents our replicated PhoNLP result. [$\blacklozenge$] represents the average results of five runs, with the lowest result indicated by [$\clubsuit$]. The red bold font indicates that our result is significantly different from the result of $\text{PhoBERT}_\text{base} [\bigstar]$ using a paired-wise t-test. POS denoted part-of-speech.\label{tab:phonlp}}
\centering
\begin{tabular}{|ll|c|c|}
\hline
\multicolumn{2}{|c|}{\textbf{Model}} & \textbf{LAS} & \textbf{UAS} \\ \hline\hline
\multicolumn{2}{|l|}{PhoNLP w/ $\text{PhoBERT}_\text{base}$ \cite{nguyen-nguyen-2021-phonlp}} & \textbf{78.17} & 84.95 \\ \hline
\multicolumn{2}{|l|}{PhoNLP w/ $\text{PhoBERT}_\text{base}$ [$\bigstar$]} & 77.38 & 84.44 \\ \hline\hline
\multicolumn{1}{|l|}{\multirow{2}{*}{\begin{tabular}[c]{@{}l@{}}W/o\\ POS tags\end{tabular}}} & HPSG w/ $\text{PhoBERT}_\text{base} [\clubsuit]$ & \textcolor{red}{\textbf{77.40}} & \textcolor{red}{\textbf{85.05}} \\ \cline{2-4} 
\multicolumn{1}{|l|}{} & HPSG w/ $\text{PhoBERT}_\text{base} [\blacklozenge]$ & 77.60 & \textbf{85.18} \\ \hline
\multicolumn{1}{|l|}{\multirow{2}{*}{\begin{tabular}[c]{@{}l@{}}W/\\ POS tags\end{tabular}}} & HPSG w/ $\text{PhoBERT}_\text{base} [\clubsuit]$ & 77.28 & \textcolor{red}{\textbf{85.01}} \\ \cline{2-4} 
\multicolumn{1}{|l|}{} & HPSG w/ $\text{PhoBERT}_\text{base} [\blacklozenge]$ & 77.48 & \textbf{85.22} \\ \hline
\end{tabular}%
\end{table}

In Table~\ref{tab:phonlp}, we report the performance of dependency parsing on the VnDT test set, highlighting the efficacy of various models using the PhoBERT base. The PhoNLP model with $\text{PhoBERT}_\text{base}$, initially achieved an LAS of 78.17 and a UAS of 84.95, while our replication of this model, denoted by [$\bigstar$], yielded slightly lower scores of 77.38 and 84.44, respectively. When examining the HPSG model with $\text{PhoBERT}_\text{base}$, both with and without POS tags, we observed varied performances. Without POS tags, the lowest scores among five runs (indicated by [$\clubsuit$]) were significantly different from the replicated PhoNLP result, as shown by the red bold font (LAS: 77.40, UAS: 85.05), while the average of these runs ([$\blacklozenge$]) showed a slightly higher LAS of 77.60 and the best UAS of 85.18. With POS tags included, the lowest and average results were 77.28 (LAS) and 85.01 (UAS) for the former, and 77.48 (LAS) and 85.22 (UAS) for the latter, with the lowest results again marked significantly different in red bold font. This analysis underscores the nuanced impact of POS tags on parsing accuracy and demonstrates the robustness of the HPSG model with $\text{PhoBERT}_\text{base}$ in dependency parsing tasks.

\subsection{Results of the VLSP 2023 Challenge on Vietnamese Constituency Parsing}

\begin{figure}[!h]
    \centering
    \begin{subfigure}[b]{0.48\linewidth}
        \includegraphics[width=\linewidth]{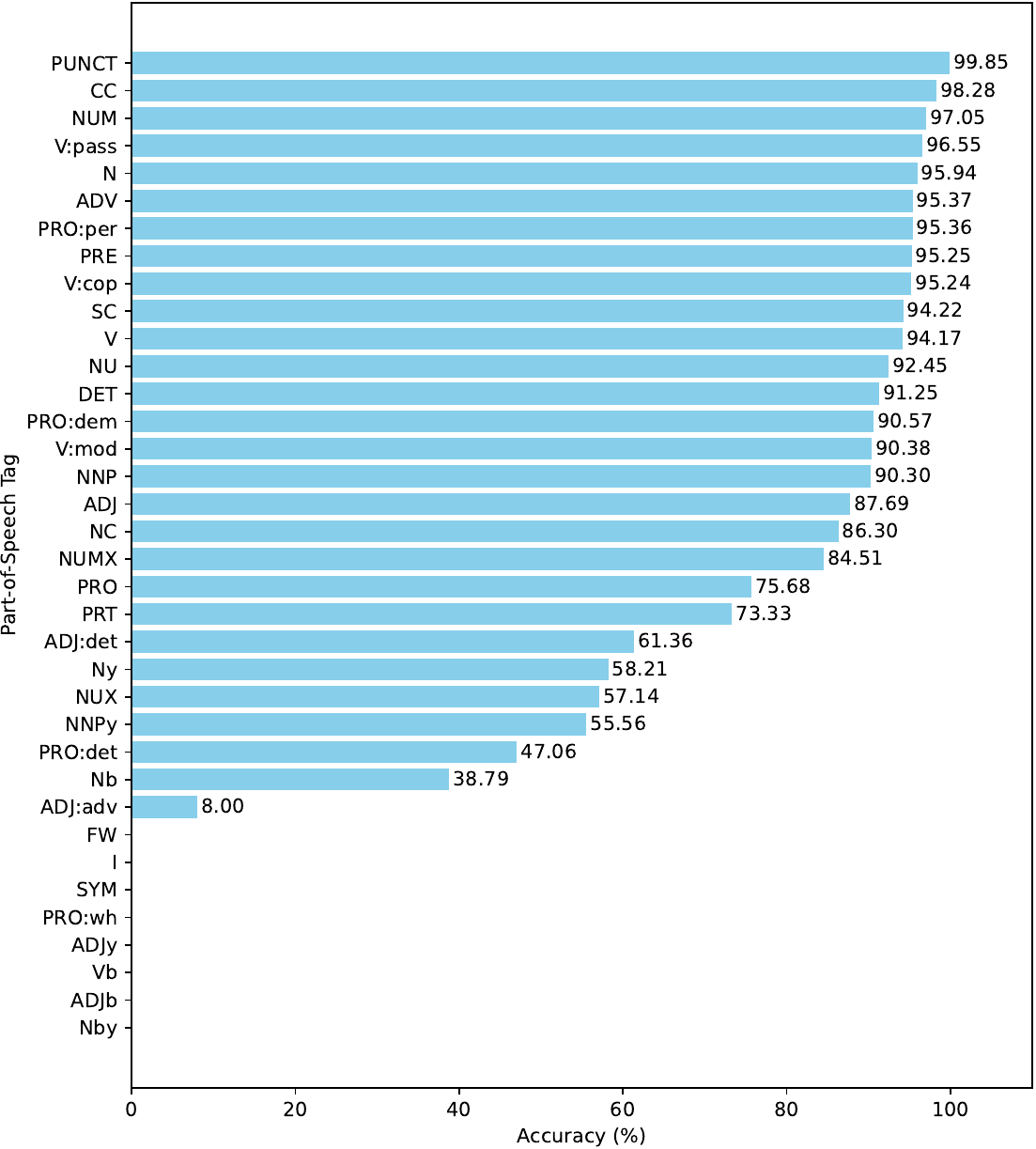}
        \caption{Tagging accuracy of different parts of speech.}
        \label{fig:tagging_detail}
    \end{subfigure}
    \hfill
    \begin{subfigure}[b]{0.48\linewidth}
        \includegraphics[width=\linewidth]{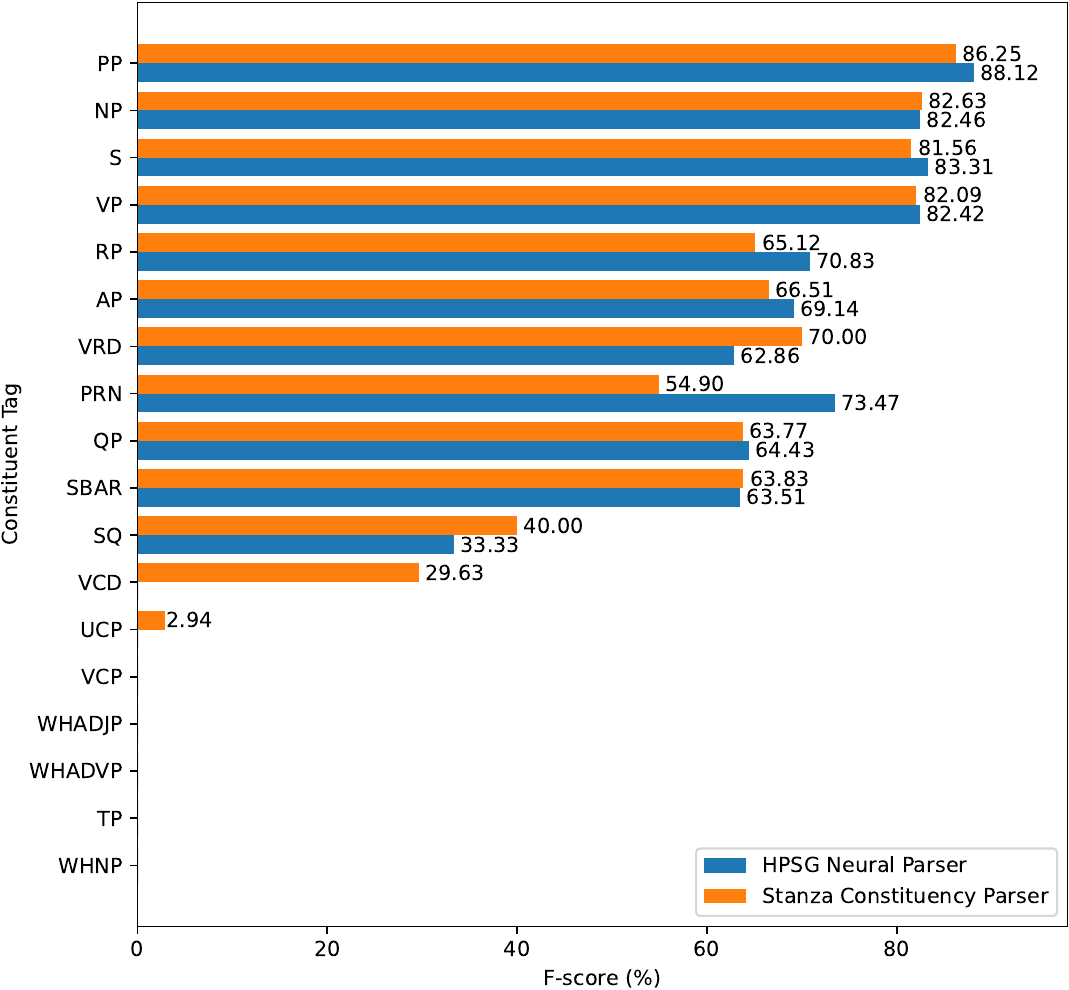}
        \caption{Parsing F-score of different constituent tags.}
        \label{fig:parsing_detail}
    \end{subfigure}
    \caption{Our tagging and parsing results on the public set of the VLSP 2023 Vietnamese Treebank.}
    \label{fig:combined_figure}
\end{figure}

The tagging results show mixed performance. We achieved very high accuracy in tagging punctuation marks (almost perfect at 0.998), coordinating conjunctions (0.983), and adverbs (0.954). This means the system is excellent at recognizing these types of words. We also got strong scores in personal pronouns (0.954), copular verbs (0.952), nouns (0.959), and verbs (0.942), indicating good performance in important word categories.

However, the system struggled with some tags. It scored zero on categories like \texttt{Nby} and \texttt{ADJb}, which means it could not identify these word types at all. It also had low scores in \texttt{PRO:det} (0.471), \texttt{Nb} (0.388), and very low in \texttt{ADJ:adv} (0.08). These are areas where we need to improve the system.

We evaluated the tagging and parsing performance on the public set of the VLSP 2023 Vietnamese Treebank. The results are presented in Figure~\ref{fig:combined_figure}.

In parsing performance, as shown in Figure~\ref{fig:combined_figure}, we provide detailed F-scores for specific syntactic categories, revealing more about each parser's strengths and weaknesses. For example, in parenthetical clauses (\texttt{PRN}), the HPSG parser scored 0.735, which is better than Stanza's 0.549. In noun phrases (\texttt{NP}), both parsers did almost the same, with Stanza slightly ahead at 0.826 compared to HPSG's 0.825.

Both parsers struggled with certain categories like \texttt{WHADVP}, \texttt{UCP}, and \texttt{VCP}, where they scored zero. This indicates these areas need more attention and improvement. By looking closely at the scores for different categories, we can better understand each parser's strengths and where they need to improve in processing natural language.

\begin{table}[!h]
\caption{Results of the VLSP 2023 Shared Task: Performance metrics of the Stanza parser \cite{Bauer_Bui_Thai_Manning_2023}, HPSG parser \cite{zhou-zhao-2019-head}, and Attach-Juxtapose parser \cite{yang2020}. Note that the result for the Attach-Juxtapose parser was reported by another team participating in the shared task. The `\&' symbol denotes an ensemble of two language models.\label{tab:main2023}}
\resizebox{\columnwidth}{!}{%
\centering
\begin{tabular}{|c|lll|lll|}
\hline
\multirow{2}{*}{\textbf{Model}} & \multicolumn{3}{c|}{\textbf{Public Test}} & \multicolumn{3}{c|}{\textbf{Private Test}} \\ \cline{2-7} 
 & \multicolumn{1}{c|}{\textbf{P}} & \multicolumn{1}{c|}{\textbf{R}} & \multicolumn{1}{c|}{\textbf{F}} & \multicolumn{1}{c|}{\textbf{P}} & \multicolumn{1}{c|}{\textbf{R}} & \multicolumn{1}{c|}{\textbf{F}} \\ \hline\hline
\multicolumn{1}{|l|}{Attach-Juxtapose w/ $\text{PhoBERT}_\text{base}$} & \multicolumn{1}{c|}{--} & \multicolumn{1}{c|}{--} & 80.55 & \multicolumn{1}{c|}{--} & \multicolumn{1}{c|}{--} & 84.66 \\ \hline
\multicolumn{1}{|l|}{Attach-Juxtapose w/ $\text{PhoBERT}_\text{base-v2}$} & \multicolumn{1}{c|}{--} & \multicolumn{1}{c|}{--} & 81.09 & \multicolumn{1}{c|}{--} & \multicolumn{1}{c|}{--} & 84.79 \\ \hline
\multicolumn{1}{|l|}{Attach-Juxtapose w/ $\text{PhoBERT}_\text{large}$} & \multicolumn{1}{c|}{--} & \multicolumn{1}{c|}{--} & 80.44 & \multicolumn{1}{c|}{--} & \multicolumn{1}{c|}{--} & 84.45 \\ \hline
\multicolumn{1}{|l|}{Attach-Juxtapose w/ [$\text{PhoBERT}_\text{base}$ \& $\text{PhoBERT}_\text{large}$]} & \multicolumn{1}{c|}{--} & \multicolumn{1}{c|}{--} & 80.87 & \multicolumn{1}{c|}{--} & \multicolumn{1}{c|}{--} & 84.60 \\ \hline
\multicolumn{1}{|l|}{Attach-Juxtapose w/ [$\text{PhoBERT}_\text{base-v2}$ \& $\text{PhoBERT}_\text{large}$]} & \multicolumn{1}{l|}{82.25} & \multicolumn{1}{l|}{79.97} & 81.09 & \multicolumn{1}{l|}{83.70} & \multicolumn{1}{l|}{86.06} & 84.86 \\ \hline\hline
\multicolumn{1}{|l|}{Stanza w/ $\text{PhoBERT}_\text{large}$} & \multicolumn{1}{l|}{86.78} & \multicolumn{1}{l|}{84.97} & 85.87 & \multicolumn{1}{l|}{89.56} & \multicolumn{1}{l|}{87.91} & 88.73 \\ \hline
\multicolumn{1}{|l|}{HPSG w/ $\text{PhoBERT}_\text{large}$} & \multicolumn{1}{l|}{86.84} & \multicolumn{1}{l|}{85.28} & 86.05 & \multicolumn{1}{l|}{89.56} & \multicolumn{1}{l|}{88.53} & 89.04 \\ \hline
\end{tabular}%
}
\end{table}

Table~\ref{tab:main2023} presents the results of the VLSP 2023 Shared Task\footnote{The VLSP 2023 Workshop Program is available at \url{https://vlsp.org.vn/vlsp2023}.}, comparing three parsers: Stanza, HPSG, and Attach-Juxtapose. The performance is measured using Precision (P), Recall (R), and F-score (F) on both public and private test datasets.

By comparing the results, we see that the Stanza and HPSG parsers perform similarly in the public test, with F-scores of 85.87 and 86.05, respectively. However, in the private test, the HPSG parser performs slightly better, achieving an F-score of 89.04 compared to Stanza's 88.73. This indicates that the HPSG parser is better at handling unseen data. The Attach-Juxtapose parser performs slightly lower than Stanza and HPSG overall but shows promising results when combined with specific models like $\text{PhoBERT}_\text{base-v2}$. This suggests that the Attach-Juxtapose parser has potential for improvement in future applications.

In summary, the results show that the HPSG parser is slightly stronger overall, especially in more challenging datasets, while Stanza remains highly competitive. The Attach-Juxtapose parser, although not the strongest here, shows room for growth with further refinements.

\section{Conclusion and Future Work}\label{conclusion}

This paper developed a neural parser for Vietnamese using a simplified Head-Driven Phrase Structure Grammar (HPSG). To address the 15\% of tree pairs in the VietTreebank and VnDT corpora that did not conform to HPSG rules, we permuted samples from the training and development sets. We then modified the original parser by incorporating PhoBERT and XLM-RoBERTa models for Vietnamese text encoding. Our experiments showed that the parser achieved an 82\% F-score in constituency parsing and outperformed previous studies in dependency parsing with a higher Unlabeled Attachment Score (UAS). The lower Labeled Attachment Score (LAS) likely resulted from not consulting linguistic experts. These results suggest the need for greater linguistic input when developing Vietnamese treebanks.

\section*{Acknowledgement}\label{conclusion}
This research is funded by University of Information Technology-Vietnam National University HoChiMinh City under grant number D1-2024-67.

\sloppy \printbibliography

\end{document}